\documentclass{article}

\usepackage{microtype}
\usepackage{graphicx}
\usepackage{subfigure}

\usepackage{booktabs} 

\usepackage{hyperref}

\usepackage[accepted]{icml2025}

\usepackage{amsmath}
\usepackage{amssymb}
\usepackage{mathtools}
\usepackage{amsthm}

\usepackage[capitalize,noabbrev]{cleveref}
\usepackage{array}
\usepackage{booktabs}
\usepackage{multirow}

\newcolumntype{C}[1]{>{\centering\arraybackslash}p{#1}}
\newcolumntype{L}[1]{>{\arraybackslash}p{#1}}
\newcolumntype{R}[1]{>{\raggedleft\arraybackslash}p{#1}}

\theoremstyle{plain}

\theoremstyle{definition}

\theoremstyle{remark}

\usepackage[textsize=tiny]{todonotes}

\icmltitlerunning{MOVE: A Mixture-of-Vision-Encoders Approach for Domain-Focused Vision-Language Processing}

\begin{document}

\twocolumn[
\icmltitle{MOVE: A Mixture-of-Vision-Encoders Approach for Domain-Focused Vision-Language Processing}



\icmlsetsymbol{equal}{*}

\begin{icmlauthorlist}
\icmlauthor{Matvey Skripkin}{comp,yyy}
\icmlauthor{Elizaveta Goncharova}{comp,hse}
\icmlauthor{Dmitrii Tarasov}{comp}
\icmlauthor{Andrey Kuznetsov}{comp,sberai}
\end{icmlauthorlist}

\icmlaffiliation{yyy}{Skolkovo Institute of Science and Technology, Moscow, Russia}
\icmlaffiliation{comp}{Artificial Intelligence Research Institute (AIRI), Moscow, Russia}
\icmlaffiliation{hse}{Higher School of Economics, Moscow, Russia}
\icmlaffiliation{sberai}{SberAI, Moscow, Russia}

\icmlcorrespondingauthor{Matvey Skripkin}{Matvey.Skripkin@airi.net}

\icmlkeywords{Machine Learning, ICML}

\vskip 0.3in
]



\printAffiliationsAndNotice{}  

\begin{abstract}
Multimodal language models (MLMs) integrate visual and textual information by coupling a vision encoder with a large language model through the specific adapter. While existing approaches commonly rely on a single pre-trained vision encoder, there is a great variability of specialized encoders that can boost model's performance in distinct domains. In this work, we propose MOVE (Mixture of Vision Encoders) a simple yet effective approach to leverage multiple pre-trained encoders for specialized multimodal tasks. MOVE automatically routes inputs to the most appropriate encoder among candidates such as Unichat, InternViT, and Texify, thereby enhancing performance across a diverse set of benchmarks, including ChartQA, MMBench, and MMMU. Experimental results demonstrate that MOVE achieves competitive accuracy without incurring the complexities of image slicing for high-resolution images.
\end{abstract}

\section{Introduction}
\label{intro}

In recent years, multimodal architectures have emerged as a powerful paradigm for advancing artificial intelligence (AI) systems, enabling them to process and understand multiple types of data simultaneously \cite{Alayrac2022FlamingoAV, Gao2023LLaMAAdapterVP, Lin2023VideoLLaVALU}. Integrating diverse data modalities, such as text and images, has significantly improved the capabilities of large language models (LLMs) in tasks ranging from visual question answering (VQA) \cite{liu2023llava} to complex decision-making \cite{Li2023LLaVAMedTA, decision_gpt}. However, effectively coupling different data types remains a major challenge for the existing multimodal models. Moreover, these multimodal multitask architectures are viewed as preliminary steps toward the development of artificial general intelligence (AGI), further expanding the scope of challenges in world cognition.

Most multimodal models rely on a single, pre-trained encoder to generate image representations, which are subsequently processed by an adapter and an LLM. While this strategy has been successful in various benchmarks, using one encoder for all domains is often insufficient. For instance, processing high-resolution images can be challenging for the vision encoders trained on fixed-resolution images. To obercome this obstacle, first, slicing techniques \cite{li2024llava} have been proposed, while the models such as LLaVA-HR \cite{luo2024feast} combine a low-resolution encoder (e.g., CLIP \cite{clip}) with a high-resolution encoder (e.g., ConvNext \cite{liu2022convnet2020s}). Additional encoders have also been introduced for grounding tasks \cite{cogvlm}, and entirely new models have been fine-tuned for specialized medical domains \cite{chexpert}. These studies indicate that domain-specific data may require specialized encoders. At the same time, such encoders can introduce extra overhead and unnecessary noise for general tasks if they are used in combination.

In this work, we propose the \emph{mixture-of-vision encoders} (MOVE) model, designed to select the most appropriate encoder for a given domain. Our approach employs a simple yet effective method: first pre-training a lightweight expert router using embeddings from a general-purpose encoder, and then inferring only the most suitable vision encoder for each input sample while using a consistent LLM. Furthermore, we show that by leveraging as few as 256, 196, or 576 vision tokens, our method attains performance comparable to state-of-the-art models that utilize slicing and substantially higher token counts.

The main contributions of our work are as follows: \textbf{(1)} We introduce a novel mixture-of-vision experts architecture that incorporates a pre-trained encoder. \textbf{(2)} We demonstrate the effectiveness of our approach across various multimodal benchmarks and LLM scales using a small number of vision tokens. \textbf{(3)} We release the source code for training and evaluating the proposed model, promoting future research and practical adoption.

\section{Related Works}

Many vision-language models extend pre-trained language models by integrating a general vision encoder, which extracts image features and aligns them with textual embeddings \cite{chen2024internvl, li2024llava}. Commonly used vision encoders include CLIP, SigLIP \cite{siglip}, InternViT, and others. These models are typically trained on large datasets of image-text pairs, allowing the extracted vision features to align naturally with textual inputs during pre-training. This alignment simplifies their integration with large language models (LLMs).

Despite this success, challenges arise in domain-specific tasks where general-purpose vision encoders may fail to capture the nuanced information necessary for strong task performance. Several studies have examined the variety of features produced by different vision encoders, highlighting notable discrepancies in their outputs. This observation has fueled efforts to design vision-language models tailored to specific domains, where encoder selection or combination is of critical importance.

Recent work has explored merging multiple vision encoders within a single multimodal architecture to improve feature representation. For instance, DeepSeek-VL employs a dual vision branch strategy: CLIP is used for semantic feature extraction, while a SAM-based encoder fine-tuned for high-resolution images captures finer details. Likewise, LLava-HR integrates high-resolution encoder outputs into CLIP’s lower-resolution feature space, thereby creating a richer representation.

The authors of the MOVA model \cite{zong2024movaadaptingmixturevision} propose a specific routing mechanism for the LLM that directs each sample to the vision encoder best suited to its domain. Their results demonstrate that each specialized encoder excels in its corresponding domain. Although effective, this approach requires two inferences of the large language model—one for selecting the appropriate encoder and another for the final inference—and relies on a complex adapter to combine the outputs from different encoders into a single LLM.

Another approach for integrating outputs from different encoders into a single embedding is presented in Eagle \cite{shi2024eagleexploringdesignspace}, where the authors investigate various fusion strategies. They conclude that dimension-based concatenation performs well across a variety of tasks without excessively expanding the context.

In more recent work \cite{azadani2025leo}, the authors propose fusing embeddings from multiple encoders to enrich the context information about an image, accounting for both high- and low-resolution features. While these studies explore different ways to merge encoders into a single embedding, in MOVE we suggest routing encoders to avoid overloading the context with unnecessary information, inferring only the one needed. Moreover, this routing mechanism is lightweight, implemented as an MLP trained on top of the universal embedding derived from CLIP.

\section{Basic Definition}
\subsection{Mixture of Encoders}

Mixture of Experts \cite{moe, fedus2022switch, zoph2022stmoedesigningstabletransferable} is actively used for Large Language models, where the feed forward layer in the attention block is replaced with the mixture-of-experts (MoE) module. The mixture of experts architecture supposes that only the specific number of the output experts (feed-forward layers) are activated for the input token $x$. Thus, the output of the MoE module is defined as follows $G(x) := \text{Softmax}(\text{TopK}(x \cdot W_g)),$ where $\text{K}$ is the number of the experts activated per token, $W_g$ is an expert parameter. The variations of several mixture of experts are modifications were introduced in \cite{jiang2024mixtralexperts, dai2024deepseekmoe}, showing the potential of applying such models.

In our work, we define experts in the more global way. Thus, the expert in MOVE is the specific vision encoder that is activated for the input sample (based on the image) through the pre-trained routing mechanism. 

\section{MOVE}

\subsection{Architecture}

\begin{figure}[ht]
\vskip 0.2in
\begin{center}
\centerline{\includegraphics[width=\columnwidth]{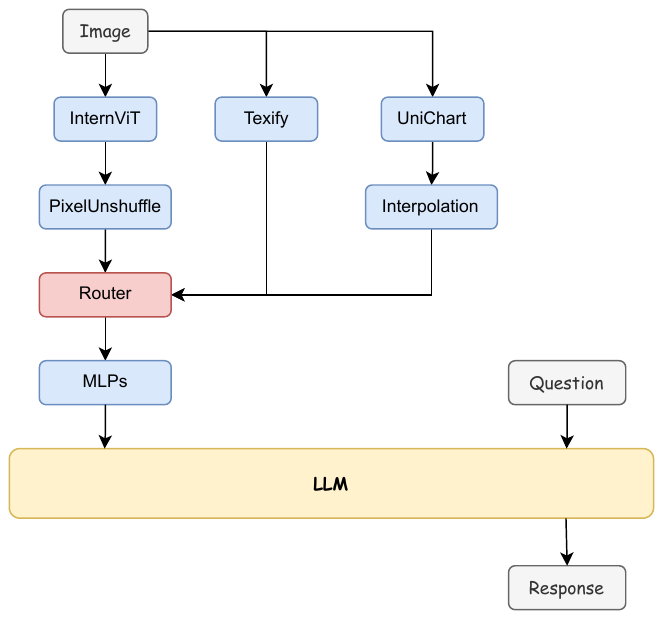}}
\caption{The MOVE architecture. The model consists of a large language model (LLM), multiple vision experts, a router for encoder selection, and adapter modules bridging visual and textual representations}
\label{fig:model_arch}
\end{center}
\vskip -0.2in
\end{figure}

Inspired by the Mixture of Experts (MoE) paradigm, we propose a modular vision-language model architecture that dynamically selects specialized vision encoders based on the input type. As illustrated in Figure~\ref{fig:model_arch}, our approach, MOVE, consists of four key components: \textbf{(1)} a pre-trained large language model (LLM), \textbf{(2)}  multiple vision expert encoders, \textbf{(3)} a router responsible for encoder selection, and \textbf{(4)} adapter modules that bridge visual and textual representations.

The integration of multiple vision encoders ensures robust feature extraction across diverse visual domains while maintaining adaptability and efficiency. In this work, we primarily use Qwen2 and Qwen2.5 (1.5B/7B) as the backbone LLMs, leveraging their advanced reasoning capabilities for multimodal processing. 

\paragraph{Pre-trained Vision Encoders}  

To effectively handle a diverse range of visual inputs, we integrate multiple state-of-the-art pre-trained vision encoders tailored to different domains: InternViT-300M-448px-v2.5 \cite{chen2024internvl}, Texify\footnote{\href{https://github.com/VikParuchuri/texify}{https://github.com/VikParuchuri/texify}}, and UniChart \cite{masry2023unichart}. InternViT specializes in natural images, Texify focuses on document analysis and LaTeX formulas, and UniChart targets structured data (e.g., charts and graphs). Each encoder delivers rich feature representations that support robust performance on downstream tasks.

\subsection{Model Architecture and Encoder Selection}  

Given an input image \( I \in \mathbb{R}^{H \times W \times C} \), MOVE follows a structured multi-stage processing pipeline:

\begin{enumerate}
    \item The image is first processed by a router, which selects the most appropriate vision encoder. 
    \item The selected encoder extracts relevant feature representations.  
    \item A corresponding adapter module transforms these features into a compatible format for the LLM. 
    \item The processed visual features are combined with textual embeddings and fed into the LLM for joint reasoning.
\end{enumerate}

\paragraph{Vision Encoding and Feature Extraction}  

Each input image \( I \) is passed through a selected vision encoder \( E_i \) from a predefined set of \( N \) encoders. The output of \( E_i \) is a feature representation:  
\[
F_i \in \mathbb{R}^{T_i \times D_i},
\]  
where \( T_i \) denotes the number of visual tokens, and \( D_i \) is the dimensionality of the extracted features. These tokens serve as the basis for downstream multimodal integration.  

\paragraph{Encoder Selection via Router}  
To determine the optimal encoder for a given input, we utilize a routing mechanism based on InternViT features. Specifically, we compute a mean-pooled feature vector:  
\[
F_{\text{avg}} = \frac{1}{T_{\text{InternViT}}} \sum_{j=1}^{T_{\text{InternViT}}} F_{\text{InternViT}}[j] \in \mathbb{R}^{1 \times D_{\text{InternViT}}}.
\]  
This vector is then processed by a linear transformation:  
\[
R(F_{\text{avg}}) = F_{\text{avg}} W, \quad W \in \mathbb{R}^{D_{\text{InternViT}} \times N}.
\]  
The router assigns the input to the most suitable encoder based on its characteristics:  
\[
F \rightarrow  
\begin{cases} 
E_1(I) = \text{InternViT}, & \text{if } I \text{ is an image-text pair}, \\
E_2(I) = \text{Texify}, & \text{if } I \text{ is a LaTeX document}, \\
E_3(I) = \text{UniChart}, & \text{if } I \text{ is a graph or chart}.
\end{cases}
\]  

InternViT is a lightweight 300M-parameter encoder whose inference time is negligible compared to that of a multi-billion-parameter LLM. Consequently, our routing scheme requires minimal resources, as it relies only on the inference of this smaller vision model.

\paragraph{Adapter and LLM Processing}  
Before projecting visual embeddings into the LLM’s embedding space, we apply preprocessing steps tailored to each vision encoder. For InternViT, we perform pixel unshuffling \cite{shi2016real} to reduce the number of output tokens, ensuring a more compact representation. For the UniChart encoder, we apply bilinear interpolation to adjust the feature map resolution before projection.  

Each vision encoder \( E_i \) has a dedicated adapter module \( A_i \) that projects its output into the LLM’s embedding space:  
\[
F_{\text{adapted}} = A_i(F_i), \quad A_i: \mathbb{R}^{T_i \times D_i} \to \mathbb{R}^{T_i \times D_{\text{LLM}}}.
\]  
These adapted visual features are then concatenated with the corresponding textual embeddings and processed by the LLM, enabling multimodal reasoning and generation.

\section{Experiments}
\label{sec:exp}

\subsection{Training Pipeline}

\begin{figure}[ht]
\vskip 0.2in
\begin{center}
\centerline{\includegraphics[width=0.3\columnwidth]{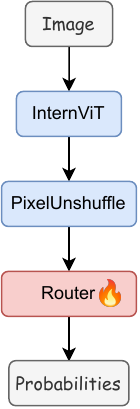}}
\caption{Router pre-training stage}
\label{fig:router}
\end{center}
\vskip -0.2in
\end{figure}

\begin{figure}[ht]
    \vskip 0.2in
    \begin{center}
    \centerline{\includegraphics[width=\columnwidth]{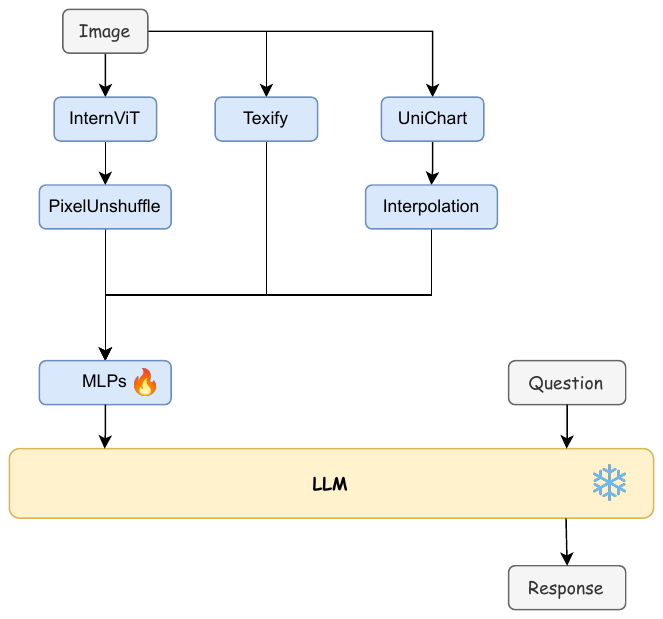}}
    \caption{Pre-training stage of the MOVE}
    \label{fig:mlp_pre-training}
    \end{center}
    \vskip -0.2in
\end{figure}

\begin{figure}[ht]
\vskip 0.2in
\begin{center}
\centerline{\includegraphics[width=\columnwidth]{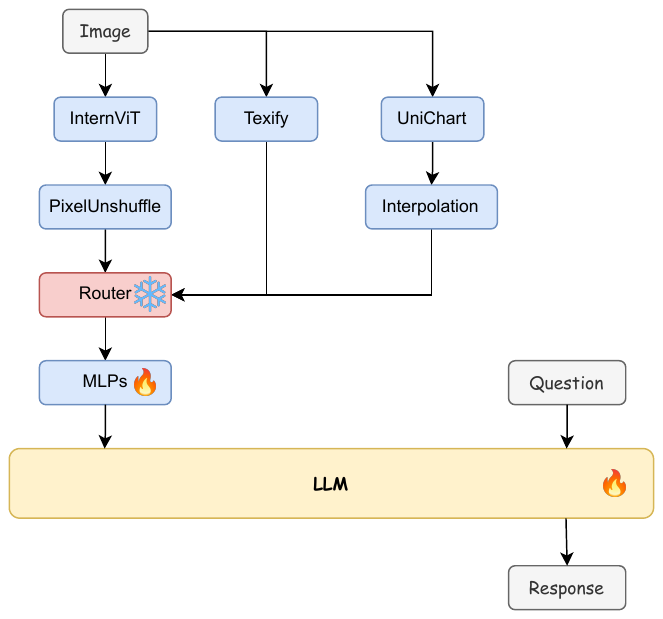}}
\caption{Supervised fine-tuning stage of the MOVE}
\label{fig:sft}
\end{center}
\vskip -0.2in
\end{figure}

We conduct training in three main stages: (1) Pre-training the router (see Figure~\ref{fig:router}) (2) Pre-training the multimodal model with both the router and the adapter (see Figure~\ref{fig:mlp_pre-training}), and (3) Fine-tuning the multimodal model with a trainable adapter and a Large Language Model (see Figure~\ref{fig:sft}).

The corresponding training parameters for each stage are listed in Table~\ref{table-params}.

\subsection{Pre-Training Stage}  

The pre-training phase aimed at learning the adapter modules and training the router for vision encoder selection. Specifically, we optimized:  

\begin{enumerate}
    \item A set of adapter modules, implemented as MLPs, each tailored to a specific vision expert. 
    \item A router, trained as a classifier to assign inputs to the appropriate vision encoder.  
\end{enumerate}

\paragraph{Adapter pre-training}  

Each adapter module \( A_i \) was trained to align the features extracted by its corresponding vision encoder \( E_i \) with the embedding space of the large language model (LLM). The following datasets were used for adapter pre-training: 

\begin{itemize}
    \item \textbf{Latex formulas (TexTeller)}\footnote{\href{https://github.com/OleehyO/TexTeller/tree/main}{https://github.com/OleehyO/TexTeller/tree/main}} – first 500k samples are taken as train set.
    \item \textbf{ShareCaptioner mix} – a diverse multimodal dataset.  
    \item \textbf{MMC-Instruct} – an instruction-following dataset for chart understanding multimodal tasks.
\end{itemize}

The datasets utilized during pre-training stage capture different captioning datasets combined with OCR-based samples.

The mix of data includes captioning including filtered Laion-EN \cite{schuhmann2022laion}, and COYO \cite{lu2023delvingdeeperdatascaling}. OCR-based part of the dataset is obtained from the LLaVAR \cite{zhang2024llavarenhancedvisualinstruction} dataset specifically collected and synthesized boosting model's abilities in OCR tasks. Finally, we included MMC-Instruct \cite{liu-etal-2024-mmc} a dataset for document-based OCR tasks captivated in instruct-based format to provide more robust training dictribution.

\paragraph{Router training}  

The router was trained in a self-supervised approach as a three-class classifier, determining which vision encoder should process a given input. The training data for the router was sampled from multiple datasets, with class labels assigned as follows:  

\begin{itemize}
    \item \textbf{Class 0} (General image-text pairs): 25K samples from LLaVA-ReCap \cite{li2024llavanext-ablations} .  
    \item \textbf{Class 1} (LaTeX documents): 25K samples from TexTeller LaTeX formulas. 
    \item \textbf{Class 2} (Structured data, charts, and diagrams):  
    
        - 25K samples from MMC-Instruct \cite{liu-etal-2024-mmc}.  
        
        - 25K samples from ChartQA \cite{masry-etal-2022-chartqa}.  
\end{itemize}

By training the router on these carefully curated datasets, we ensured that it effectively distinguishes between different input modalities, enabling the model to route images to the most appropriate vision encoder corresponding to some specific domain.  

\subsection{Fine-Tuning stage}  

After pre-training, we fine-tuned the multimodal model while keeping the vision experts and the router frozen. This step focused on optimizing the large language model (LLM) and the vision adapters to improve their ability to integrate visual and textual information effectively.  

For fine-tuning, we utilized the datasets given in Table \ref{tab:datasets-distrib}.

\begin{table}[ht]
\centering
\label{tab:datasets-distrib}
\caption{Datasets distribution used for training the MOVE model}
\label{tab:datasets}
\begin{tabular}{L{1.6cm} R{2.7cm} R{1.8cm}}
\toprule
\textbf{Category} & \textbf{Dataset Name} & \textbf{\# Train Samples} \\
\midrule
\multirow{7}{*}{\textbf{Graphs}}
 & AI2D                    & 2.43k    \\
 & Chart2Text             & 27k    \\
 & ChartQA                & 18.3k    \\
 & Diagram (image-to-text) & 0.3k  \\
 & dVQA                   & 50k    \\
 & FigureQA               & 50k    \\
 & Infographic            & 2.12k    \\
\midrule
\multirow{2}{*}{\textbf{Documents}}
 & docVQA                 & 10.2k    \\
 & OCRVQA                 & 50k    \\
\midrule
\multirow{7}{*}{\begin{tabular}{l}
\textbf{General}\\
\textbf{Domain}
\end{tabular}}
 & VQA                    & --    \\
 & OKVQA                  & 9.01k    \\
 & scienceQA              & 4.98k    \\
 & TextVQA                & 1.49k    \\
 & TQA                    & 50k   \\
 & Cambrian (subsample)   & 737k  \\
 & Latex formula          & 500k  \\
\bottomrule
\end{tabular}
\end{table}





\section{Implementation Details}

As we mentioned in Section \ref{sec:exp}, we implemented three-seteps training pipeline: router training, MOVE pre-training, MOVE-finetuning. The hyperparameters used during these three stages are given in Table \ref{table-params}.

\begin{table}[t]
\caption{Training parameters for each stage of the pipeline}
\label{table-params}
\vskip 0.15in
\begin{center}
\begin{small}
\begin{tabular}{m{1.9cm} m{1cm} m{1cm} m{1cm} m{1cm}}
\toprule
Stage & Learning Rate & Optimizer & Batch Size & Number of Steps \\
\midrule 
Router training & 1e-2 & AdamW & 12 & 8500 \\
MLM pre-training & 2e-3 & AdamW & 256 & 7200 \\
MLM fine-tuning & 1e-5 & AdamW & 192 & 7700 \\
\bottomrule
\end{tabular} 
\end{small} 
\end{center} 
\vskip -0.1in 
\end{table}

\subsection{Encoders Description}  

In MOVE, we utilize three different encoders: InternViT-2.5-distil, UniChart, and Texify. Table~\ref{tab:encoders} provides an overall summary of the utilized encoders.  

To optimize token efficiency, we apply preprocessing techniques tailored to each encoder. Specifically, we use the pixel unshuffling technique for the InternViT encoder, reducing the number of tokens from 1024 to 256. Similarly, for the UniChart encoder, we apply bilinear interpolation to decrease the number of tokens from 900 to 576.

\begin{table}[ht!]
\caption{Summary of the encoders parameters}
\centering
\label{tab:encoders}
\begin{tabular}{L{2.5cm} C{1.2cm} C{1.2cm} C{1.2cm}}
\toprule
\textbf{Model Name} & \textbf{Train Res.} & \textbf{Width} & \textbf{\#Param} \\
\midrule
InternViT-300M-448px-V2.5 & dynamic 448 & 1024  & 0.3B \\
UniChart & 960 & 900  & 74M \\
Texify & 420 & 196  & 74M \\
\bottomrule
\end{tabular}
\end{table}

\subsection{Evaluation Results}

We evaluated the proposed MOVE on the several multimodal benchmarks GQA \cite{hudson2019gqanewdatasetrealworld}, SQA \cite{lu2022learnexplainmultimodalreasoning}, MMMU \cite{yue2024mmmumassivemultidisciplinemultimodal}, MMStar \cite{chen2024rightwayevaluatinglarge}, MMBench \cite{liu2024mmbenchmultimodalmodelallaround} (general VQA and common knowledge), ChartQA \cite{masry-etal-2022-chartqa}, infoVQA \cite{mathew2021infographicvqa}, AI2D \cite{ai2d} (graph analysis), docVQA \cite{mathew2021docvqadatasetvqadocument} (OCR-based tasks). These benchmarks measures various multimodal abilities, from undersatanding the common scene on the image till reading and solving tasks based on the written texts. The main purpose of our evaluation was to understand the influence of routing mechanism on the performance on downstream task.

As presented in Table~\ref{tab:move-results}, MOVE achieves results comparable to the MOVA architecture of a similar scale. For instance, MOVE-7B outperforms MOVA on ChartQA (72.5 vs. 70.5). However, MOVE underperforms on DocVQA, likely because it lacks a specialized encoder for document-oriented or OCR-based inputs. This gap suggests that adding a dedicated document encoder could substantially improve performance on OCR-heavy tasks.

We also note that on general-domain benchmarks, MOVE relies on a standard InternViT model, using only a single image input (i.e., 256 visual tokens). In contrast, InternVL can slice an image into up to six segments (leading to more than 1K tokens per one image). Despite fewer visual tokens, MOVE still performs strongly overall, indicating that the routing mechanism effectively directs each image to the most relevant encoder within a relatively lightweight setup.

It is worth noting that MOVE, which uses between 196 and 576 visual tokens, strongly outperforms LLaVA-1.5 — another non-slicing model occupying 576 tokens per image—particularly on task-oriented benchmarks such as AI2D, ChartQA, and InfoVQA.

Interestingly, scaling up the language backbone from 1.5B to 7B parameters yields consistent gains across most benchmarks, pointing to the benefits of larger LLMs in multimodal settings. Overall, these findings underscore the versatility of MOVE and highlight potential avenues for improvement, such as introducing specialized encoders for OCR or other domain-specific tasks.

In Figures 5 and 6, we present qualitative examples of how MOVE processes LaTeX formulas. The model's task is to generate LaTeX code for the given input images. We compare MOVE’s output with that of LLaVA-OneVision~7B and observe a significantly closer match to the ground-truth formulas in MOVE's results.

\begin{figure}[h]
    \vskip 0.05in
    \begin{center}
    \includegraphics[width=\columnwidth]{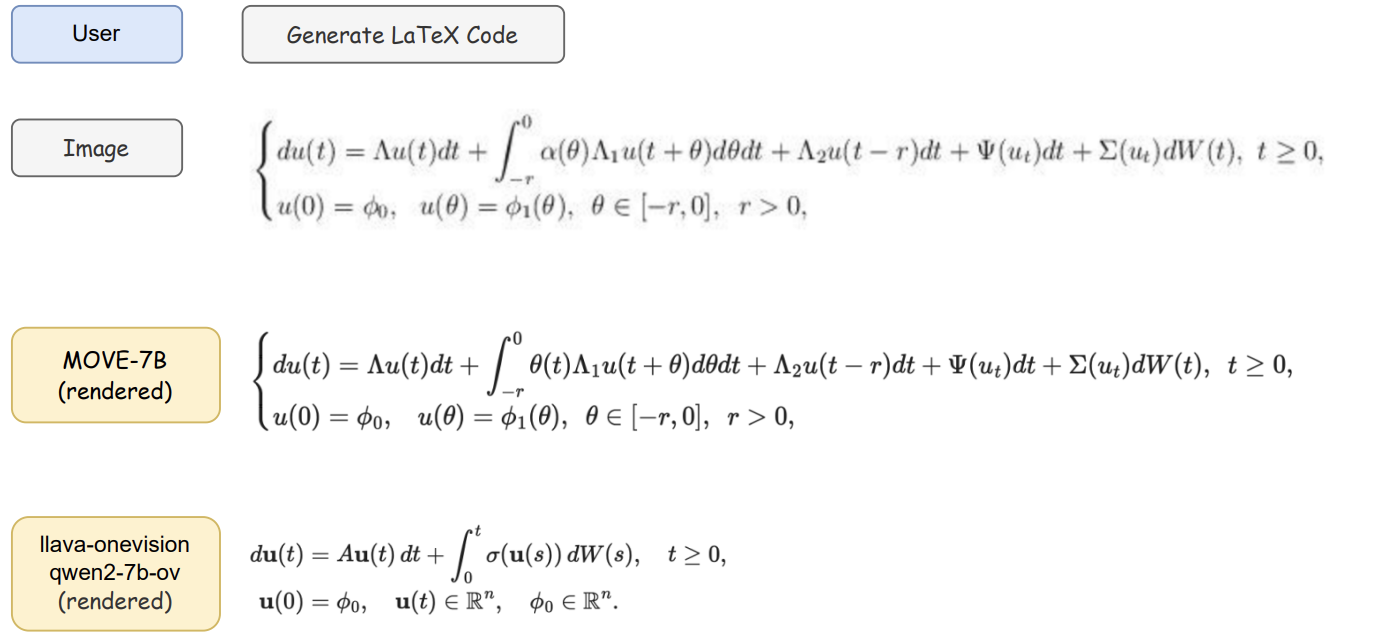}
    \label{fig:latex-1}
    \caption{Example highlighting MOVE's consistency with the ground truth in LaTeX code generation, compared to LLaVA-OneVision 7B}
    \end{center}
    \vskip -0.05in
\end{figure}

\begin{figure}[h]
    \centering
    \includegraphics[width=\columnwidth]{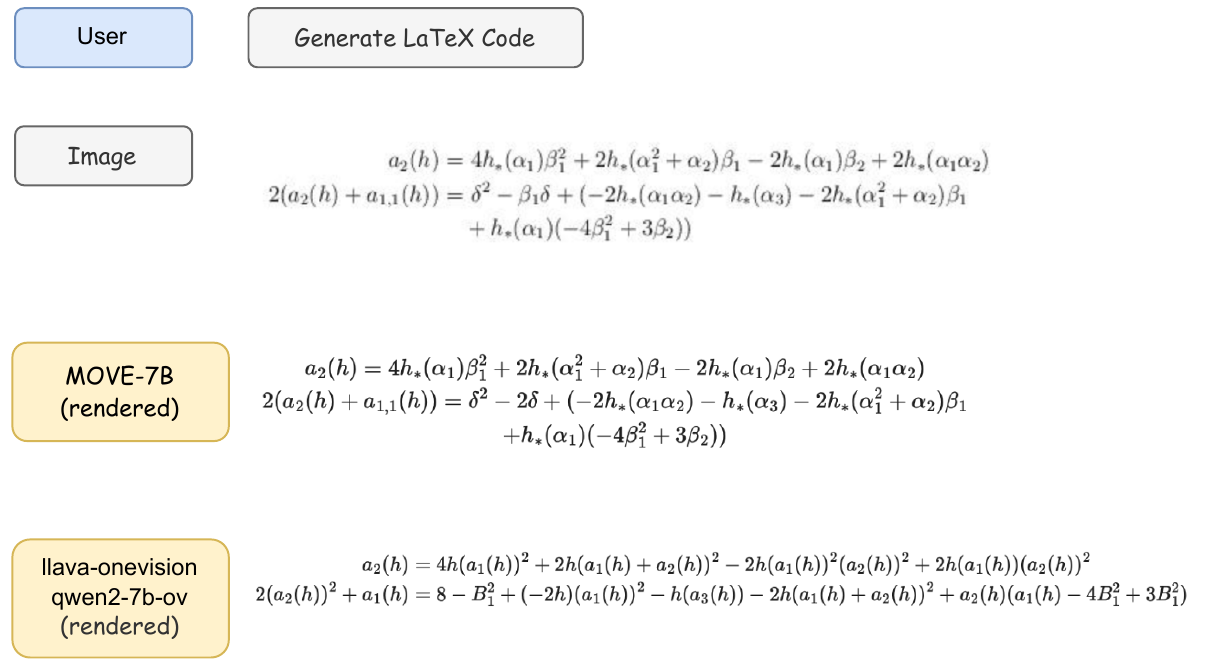}
    \caption{Example highlighting MOVE's consistency with the ground truth in LaTeX code generation, compared to LLaVA-OneVision 7B}
    \label{fig:latex-2}
\end{figure}

\begin{table*}[t!]
\caption{MOVE performance across multiple multimodal benchmarks, grouped by domain}
\label{tab:move-results}
\centering
\begin{small}
\resizebox{\textwidth}{!}{%
\begin{tabular}{%
  L{1.6cm}  
  L{1.2cm}  
  R{1.2cm}  
  R{1.2cm}  
  R{1.2cm}  
  R{1.2cm}  
  R{1.2cm}  
  R{1.2cm}  
  R{1.2cm}  
  R{0.8cm}  
  R{1.0cm}  
}
\toprule
& &
\multicolumn{3}{c}{\textbf{Graph Analysis}} &
\multicolumn{5}{c}{\textbf{General VQA \& Knowledge}} &
\multicolumn{1}{c}{\textbf{OCR}} \\
\cmidrule(lr){3-5}\cmidrule(lr){6-10}\cmidrule(lr){11-11}
\textbf{Model} & \textbf{Model size} & 
\textbf{AI2D} & \textbf{ChartQA} & \textbf{infoVQA} & 
\textbf{GQA} & \textbf{MMBench} & \textbf{MMMU} & \textbf{MMStar} & \textbf{SQA\textsubscript{img}} &
\textbf{docVQA} \\
\midrule
\multicolumn{11}{c}{\textbf{Slicing (up to 6x) for high-resolution images}} \\
\midrule
\textbf{InternVL} & InternLM2-1.8B
& 67.6 & 74.8 & 53.1 & 59.9 & 72.5 & 34.6 & 48.4 & 94.1 & 84.9 \\
\textbf{InternVL} & Llama3-8B
& \textbf{80.0*} & \textbf{82.7*} & \textbf{66.8*} & 62.8 & \textbf{81.6*} & 47.7 & 59.2 & 90.1 & 77.0 \\
\midrule
\textbf{LLaVA OneVision} & Qwen2.5-0.5B
& 54.6 & 61.4 & 40.1 & 57.8 & 51.7 & 31.9 & 39.3 & 59.3 & 68.8 \\
\textbf{LLaVA OneVision} & Qwen2-7B
& 78.8 & 79.9 & 66.1 & 62.2 & 80.8 & \textbf{49.1*} & \textbf{62.3*} & \textbf{96.0*} & \textbf{87.1*} \\
\midrule
\multicolumn{11}{c}{\textbf{No slicing for high-resolution images}} \\
\midrule
\textbf{MOVA (576 tokens)} & Vicuna-7B
& 74.9 & 68.3 & -- & 64.8 & -- & -- & -- & 81.3 & 76.4 \\
\textbf{MOVA (576 tokens)} & Llama3-8B
& \textbf{77.0} & 70.5 & -- & \textbf{65.2} & -- & -- & -- & \textbf{83.4} & 77.1 \\
\midrule
\textbf{LLaVA-1.5 (576 tokens)} & Vicuna-7B
& 57.8 & 18.0 & 20.2 & 62.0 & 63.9 & 36.3 & 33.2 & 69.3 & 21.5 \\
\midrule
\textbf{MOVE (196/256/576 tokens)} & Qwen2.5-1.5B
& 65.0 & 61.0 & 30.0 & 58.0 & 65.0 & 40.0 & 37.0 & 75.0 & 51.0 \\
\textbf{MOVE (196/256/576 tokens)} & Qwen2.5-7B
& 71.0 & \textbf{72.5} & 36.0 & 61.0 & \textbf{75.0} & \textbf{47.6} & \textbf{44.0} & 76.0 & 56.0 \\
\bottomrule
\end{tabular}
}
\end{small}
\end{table*}

\section{Conclusion}

In this work, we introduced \textbf{MOVE}, a novel multimodal architecture that enables efficient inference across various domains. Our approach relies on a pre-trained light-weight router mechanism that selects the most promising vision expert based on the classification of image tokens. We demonstrate that, even without using slicing techniques to process high-resolution images, MOVE achieves strong results on OCR-based benchmarks while utilizing significantly fewer tokens for vision input.

The proposed technique supports a wide range of data domains — such as charts, LaTeX formulas, and general images — while keeping the model compact and effective. In future research, we plan to explore end-to-end training of both the router and the multimodal components. We also aim to expand the selection of vision encoders to include additional domains, such as document-based OCR tasks (e.g., Pix2Struct~\cite{lee2023pix2structscreenshotparsingpretraining}) and medical imaging (e.g., CheXPert~\cite{chexpert}), further broadening the versatility of our approach.

\section*{Impact Statement}

This paper presents work whose goal is to advance the field of Machine Learning-specifically through the development of multimodal models capable of processing and integrating information from multiple data sources (e.g., text and images). There are many potential societal consequences of our work, none which we feel must be specifically highlighted here. However, we acknowledge that multimodal systems could impact society in several ways:

Bias and Fairness

Multimodal datasets may reflect biases present in their constituent modalities (e.g., linguistic or visual bias). Although we used open-sourced and carefully selected and processed data, there is a risk that unintended biases remain, potentially affecting system outputs in ways that disadvantage certain groups.

Potential Misuse

As models become more powerful and general-purpose, they might be leveraged for malicious applications (e.g., disinformation, surveillance). While our research focuses on advancing technical capability, we encourage ethical implementation aligned with responsible-use principles to avoid harmful outcomes.

Interpretability and Accountability

Multimodal models can be complex, making them difficult to interpret or explain. This may pose challenges in high-stakes domains (e.g., medicine, law), where accountability and transparency are important. We highlight the need for further research into explainability and clear guidelines around system deployment.

Despite these considerations, our primary aim is scientific advancement, and the benefits of better multimodal analysis. Through this work, we hope to contribute positively to the growing body of knowledge in Machine Learning and encourage ongoing dialogue about potential ethical and societal implications.


\bibliographystyle{icml2025}
\bibliography{example_paper}

\begin{thebibliography}{39}
\providecommand{\natexlab}[1]{#1}
\providecommand{\url}[1]{\texttt{#1}}
\expandafter\ifx\csname urlstyle\endcsname\relax
  \providecommand{\doi}[1]{doi: #1}\else
  \providecommand{\doi}{doi: \begingroup \urlstyle{rm}\Url}\fi

\bibitem[Alayrac et~al.(2022)Alayrac, Donahue, Luc, Miech, Barr, Hasson, Lenc, Mensch, Millican, Reynolds, Ring, Rutherford, Cabi, Han, Gong, Samangooei, Monteiro, Menick, Borgeaud, Brock, Nematzadeh, Sharifzadeh, Binkowski, Barreira, Vinyals, Zisserman, and Simonyan]{Alayrac2022FlamingoAV}
Jean-Baptiste Alayrac, Jeff Donahue, Pauline Luc, Antoine Miech, Iain Barr, Yana Hasson, Karel Lenc, Arthur Mensch, Katie Millican, Malcolm Reynolds, Roman Ring, Eliza Rutherford, Serkan Cabi, Tengda Han, Zhitao Gong, Sina Samangooei, Marianne Monteiro, Jacob Menick, Sebastian Borgeaud, Andy Brock, Aida Nematzadeh, Sahand Sharifzadeh, Mikolaj Binkowski, Ricardo Barreira, Oriol Vinyals, Andrew Zisserman, and Karen Simonyan.
\newblock Flamingo: a visual language model for few-shot learning.
\newblock \emph{ArXiv}, abs/2204.14198, 2022.
\newblock URL \url{https://api.semanticscholar.org/CorpusID:248476411}.

\bibitem[Azadani et~al.(2025)Azadani, Riddell, Sedwards, and Czarnecki]{azadani2025leo}
Mozhgan~Nasr Azadani, James Riddell, Sean Sedwards, and Krzysztof Czarnecki.
\newblock Leo: Boosting mixture of vision encoders for multimodal large language models.
\newblock \emph{arXiv preprint arXiv:2501.06986}, 2025.

\bibitem[Chen et~al.(2023)Chen, Zhang, Ren, Zhao, Cai, Wang, Wang, Liu, and Chang]{decision_gpt}
Liang Chen, Yichi Zhang, Shuhuai Ren, Haozhe Zhao, Zefan Cai, Yuchi Wang, Peiyi Wang, Tianyu Liu, and Baobao Chang.
\newblock Towards end-to-end embodied decision making via multi-modal large language model: Explorations with gpt4-vision and beyond.
\newblock \emph{CoRR}, abs/2310.02071, 2023.
\newblock \doi{10.48550/ARXIV.2310.02071}.
\newblock URL \url{https://doi.org/10.48550/arXiv.2310.02071}.

\bibitem[Chen et~al.(2024{\natexlab{a}})Chen, Li, Dong, Zhang, Zang, Chen, Duan, Wang, Qiao, Lin, and Zhao]{chen2024rightwayevaluatinglarge}
Lin Chen, Jinsong Li, Xiaoyi Dong, Pan Zhang, Yuhang Zang, Zehui Chen, Haodong Duan, Jiaqi Wang, Yu~Qiao, Dahua Lin, and Feng Zhao.
\newblock Are we on the right way for evaluating large vision-language models?, 2024{\natexlab{a}}.
\newblock URL \url{https://arxiv.org/abs/2403.20330}.

\bibitem[Chen et~al.(2024{\natexlab{b}})Chen, Wu, Wang, Su, Chen, Xing, Zhong, Zhang, Zhu, Lu, et~al.]{chen2024internvl}
Zhe Chen, Jiannan Wu, Wenhai Wang, Weijie Su, Guo Chen, Sen Xing, Muyan Zhong, Qinglong Zhang, Xizhou Zhu, Lewei Lu, et~al.
\newblock Internvl: Scaling up vision foundation models and aligning for generic visual-linguistic tasks.
\newblock In \emph{Proceedings of the IEEE/CVF Conference on Computer Vision and Pattern Recognition}, pp.\  24185--24198, 2024{\natexlab{b}}.

\bibitem[Dai et~al.(2024)Dai, Deng, Zhao, Xu, Gao, Chen, Li, Zeng, Yu, Wu, Xie, Li, Huang, Luo, Ruan, Sui, and Liang]{dai2024deepseekmoe}
Damai Dai, Chengqi Deng, Chenggang Zhao, R.~X. Xu, Huazuo Gao, Deli Chen, Jiashi Li, Wangding Zeng, Xingkai Yu, Y.~Wu, Zhenda Xie, Y.~K. Li, Panpan Huang, Fuli Luo, Chong Ruan, Zhifang Sui, and Wenfeng Liang.
\newblock Deepseekmoe: Towards ultimate expert specialization in mixture-of-experts language models, 2024.
\newblock URL \url{https://arxiv.org/abs/2401.06066}.

\bibitem[Fedus et~al.(2022)Fedus, Zoph, and Shazeer]{fedus2022switch}
William Fedus, Barret Zoph, and Noam Shazeer.
\newblock Switch transformers: Scaling to trillion parameter models with simple and efficient sparsity, 2022.

\bibitem[Gao et~al.(2023)Gao, Han, Zhang, Lin, Geng, Zhou, Zhang, Lu, He, Yue, Li, and Qiao]{Gao2023LLaMAAdapterVP}
Peng Gao, Jiaming Han, Renrui Zhang, Ziyi Lin, Shijie Geng, Aojun Zhou, W.~Zhang, Pan Lu, Conghui He, Xiangyu Yue, Hongsheng Li, and Yu~Jiao Qiao.
\newblock Llama-adapter v2: Parameter-efficient visual instruction model.
\newblock \emph{ArXiv}, abs/2304.15010, 2023.
\newblock URL \url{https://api.semanticscholar.org/CorpusID:258418343}.

\bibitem[Hudson \& Manning(2019)Hudson and Manning]{hudson2019gqanewdatasetrealworld}
Drew~A. Hudson and Christopher~D. Manning.
\newblock Gqa: A new dataset for real-world visual reasoning and compositional question answering, 2019.
\newblock URL \url{https://arxiv.org/abs/1902.09506}.

\bibitem[Irvin et~al.(2019)Irvin, Rajpurkar, Ko, Yu, Ciurea-Ilcus, Chute, Marklund, Haghgoo, Ball, Shpanskaya, Seekins, Mong, Halabi, Sandberg, Jones, Larson, Langlotz, Patel, Lungren, and Ng]{chexpert}
Jeremy Irvin, Pranav Rajpurkar, Michael Ko, Yifan Yu, Silviana Ciurea-Ilcus, Chris Chute, Henrik Marklund, Behzad Haghgoo, Robyn Ball, Katie Shpanskaya, Jayne Seekins, David~A. Mong, Safwan~S. Halabi, Jesse~K. Sandberg, Ricky Jones, David~B. Larson, Curtis~P. Langlotz, Bhavik~N. Patel, Matthew~P. Lungren, and Andrew~Y. Ng.
\newblock Chexpert: a large chest radiograph dataset with uncertainty labels and expert comparison.
\newblock In \emph{Proceedings of the Thirty-Third AAAI Conference on Artificial Intelligence and Thirty-First Innovative Applications of Artificial Intelligence Conference and Ninth AAAI Symposium on Educational Advances in Artificial Intelligence}, AAAI'19/IAAI'19/EAAI'19. AAAI Press, 2019.
\newblock ISBN 978-1-57735-809-1.
\newblock \doi{10.1609/aaai.v33i01.3301590}.
\newblock URL \url{https://doi.org/10.1609/aaai.v33i01.3301590}.

\bibitem[Jiang et~al.(2024)Jiang, Sablayrolles, Roux, Mensch, Savary, Bamford, Chaplot, de~las Casas, Hanna, Bressand, Lengyel, Bour, Lample, Lavaud, Saulnier, Lachaux, Stock, Subramanian, Yang, Antoniak, Scao, Gervet, Lavril, Wang, Lacroix, and Sayed]{jiang2024mixtralexperts}
Albert~Q. Jiang, Alexandre Sablayrolles, Antoine Roux, Arthur Mensch, Blanche Savary, Chris Bamford, Devendra~Singh Chaplot, Diego de~las Casas, Emma~Bou Hanna, Florian Bressand, Gianna Lengyel, Guillaume Bour, Guillaume Lample, Lélio~Renard Lavaud, Lucile Saulnier, Marie-Anne Lachaux, Pierre Stock, Sandeep Subramanian, Sophia Yang, Szymon Antoniak, Teven~Le Scao, Théophile Gervet, Thibaut Lavril, Thomas Wang, Timothée Lacroix, and William~El Sayed.
\newblock Mixtral of experts, 2024.
\newblock URL \url{https://arxiv.org/abs/2401.04088}.

\bibitem[Kembhavi et~al.(2016)Kembhavi, Salvato, Kolve, Seo, Hajishirzi, and Farhadi]{ai2d}
Aniruddha Kembhavi, Mike Salvato, Eric Kolve, Minjoon Seo, Hannaneh Hajishirzi, and Ali Farhadi.
\newblock A diagram is worth a dozen images.
\newblock In Bastian Leibe, Jiri Matas, Nicu Sebe, and Max Welling (eds.), \emph{Computer Vision -- ECCV 2016}, pp.\  235--251, Cham, 2016. Springer International Publishing.
\newblock ISBN 978-3-319-46493-0.

\bibitem[Lee et~al.(2023)Lee, Joshi, Turc, Hu, Liu, Eisenschlos, Khandelwal, Shaw, Chang, and Toutanova]{lee2023pix2structscreenshotparsingpretraining}
Kenton Lee, Mandar Joshi, Iulia Turc, Hexiang Hu, Fangyu Liu, Julian Eisenschlos, Urvashi Khandelwal, Peter Shaw, Ming-Wei Chang, and Kristina Toutanova.
\newblock Pix2struct: Screenshot parsing as pretraining for visual language understanding, 2023.
\newblock URL \url{https://arxiv.org/abs/2210.03347}.

\bibitem[Li et~al.(2024{\natexlab{a}})Li, Zhang, Zhang, Guo, Zhang, Zhang, Li, Liu, and Li]{li2024llavanext-ablations}
Bo~Li, Hao Zhang, Kaichen Zhang, Dong Guo, Yuanhan Zhang, Renrui Zhang, Feng Li, Ziwei Liu, and Chunyuan Li.
\newblock Llava-next: What else influences visual instruction tuning beyond data?, May 2024{\natexlab{a}}.
\newblock URL \url{https://llava-vl.github.io/blog/2024-05-25-llava-next-ablations/}.

\bibitem[Li et~al.(2023)Li, Wong, Zhang, Usuyama, Liu, Yang, Naumann, Poon, and Gao]{Li2023LLaVAMedTA}
Chunyuan Li, Cliff Wong, Sheng Zhang, Naoto Usuyama, Haotian Liu, Jianwei Yang, Tristan Naumann, Hoifung Poon, and Jianfeng Gao.
\newblock Llava-med: Training a large language-and-vision assistant for biomedicine in one day.
\newblock \emph{ArXiv}, abs/2306.00890, 2023.
\newblock URL \url{https://api.semanticscholar.org/CorpusID:258999820}.

\bibitem[Li et~al.(2024{\natexlab{b}})Li, Zhang, Zhang, Zhang, Li, Li, Ma, and Li]{li2024llava}
Feng Li, Renrui Zhang, Hao Zhang, Yuanhan Zhang, Bo~Li, Wei Li, Zejun Ma, and Chunyuan Li.
\newblock Llava-next-interleave: Tackling multi-image, video, and 3d in large multimodal models.
\newblock \emph{arXiv preprint arXiv:2407.07895}, 2024{\natexlab{b}}.

\bibitem[Lin et~al.(2023)Lin, Zhu, Ye, Ning, Jin, and Yuan]{Lin2023VideoLLaVALU}
Bin Lin, Bin Zhu, Yang Ye, Munan Ning, Peng Jin, and Li~Yuan.
\newblock Video-llava: Learning united visual representation by alignment before projection.
\newblock In \emph{Conference on Empirical Methods in Natural Language Processing}, 2023.
\newblock URL \url{https://api.semanticscholar.org/CorpusID:265281544}.

\bibitem[Liu et~al.(2024{\natexlab{a}})Liu, Wang, Yao, Chen, Song, Cho, Yacoob, and Yu]{liu-etal-2024-mmc}
Fuxiao Liu, Xiaoyang Wang, Wenlin Yao, Jianshu Chen, Kaiqiang Song, Sangwoo Cho, Yaser Yacoob, and Dong Yu.
\newblock {MMC}: Advancing multimodal chart understanding with large-scale instruction tuning.
\newblock In Kevin Duh, Helena Gomez, and Steven Bethard (eds.), \emph{Proceedings of the 2024 Conference of the North American Chapter of the Association for Computational Linguistics: Human Language Technologies (Volume 1: Long Papers)}, pp.\  1287--1310, Mexico City, Mexico, June 2024{\natexlab{a}}. Association for Computational Linguistics.
\newblock \doi{10.18653/v1/2024.naacl-long.70}.
\newblock URL \url{https://aclanthology.org/2024.naacl-long.70/}.

\bibitem[Liu et~al.(2023)Liu, Li, Wu, and Lee]{liu2023llava}
Haotian Liu, Chunyuan Li, Qingyang Wu, and Yong~Jae Lee.
\newblock Visual instruction tuning.
\newblock In \emph{NeurIPS}, 2023.

\bibitem[Liu et~al.(2024{\natexlab{b}})Liu, Duan, Zhang, Li, Zhang, Zhao, Yuan, Wang, He, Liu, Chen, and Lin]{liu2024mmbenchmultimodalmodelallaround}
Yuan Liu, Haodong Duan, Yuanhan Zhang, Bo~Li, Songyang Zhang, Wangbo Zhao, Yike Yuan, Jiaqi Wang, Conghui He, Ziwei Liu, Kai Chen, and Dahua Lin.
\newblock Mmbench: Is your multi-modal model an all-around player?, 2024{\natexlab{b}}.
\newblock URL \url{https://arxiv.org/abs/2307.06281}.

\bibitem[Liu et~al.(2022)Liu, Mao, Wu, Feichtenhofer, Darrell, and Xie]{liu2022convnet2020s}
Zhuang Liu, Hanzi Mao, Chao-Yuan Wu, Christoph Feichtenhofer, Trevor Darrell, and Saining Xie.
\newblock A convnet for the 2020s, 2022.
\newblock URL \url{https://arxiv.org/abs/2201.03545}.

\bibitem[Lu et~al.(2023)Lu, Jin, Hou, Liew, Cheng, and Feng]{lu2023delvingdeeperdatascaling}
Cheng-Ze Lu, Xiaojie Jin, Qibin Hou, Jun~Hao Liew, Ming-Ming Cheng, and Jiashi Feng.
\newblock Delving deeper into data scaling in masked image modeling, 2023.
\newblock URL \url{https://arxiv.org/abs/2305.15248}.

\bibitem[Lu et~al.(2022)Lu, Mishra, Xia, Qiu, Chang, Zhu, Tafjord, Clark, and Kalyan]{lu2022learnexplainmultimodalreasoning}
Pan Lu, Swaroop Mishra, Tony Xia, Liang Qiu, Kai-Wei Chang, Song-Chun Zhu, Oyvind Tafjord, Peter Clark, and Ashwin Kalyan.
\newblock Learn to explain: Multimodal reasoning via thought chains for science question answering, 2022.
\newblock URL \url{https://arxiv.org/abs/2209.09513}.

\bibitem[Luo et~al.(2024)Luo, Zhou, Zhang, Zheng, Sun, and Ji]{luo2024feast}
Gen Luo, Yiyi Zhou, Yuxin Zhang, Xiawu Zheng, Xiaoshuai Sun, and Rongrong Ji.
\newblock Feast your eyes: Mixture-of-resolution adaptation for multimodal large language models.
\newblock \emph{arXiv preprint arXiv:2403.03003}, 2024.

\bibitem[Masry et~al.(2022)Masry, Do, Tan, Joty, and Hoque]{masry-etal-2022-chartqa}
Ahmed Masry, Xuan~Long Do, Jia~Qing Tan, Shafiq Joty, and Enamul Hoque.
\newblock {C}hart{QA}: A benchmark for question answering about charts with visual and logical reasoning.
\newblock In Smaranda Muresan, Preslav Nakov, and Aline Villavicencio (eds.), \emph{Findings of the Association for Computational Linguistics: ACL 2022}, pp.\  2263--2279, Dublin, Ireland, May 2022. Association for Computational Linguistics.
\newblock \doi{10.18653/v1/2022.findings-acl.177}.
\newblock URL \url{https://aclanthology.org/2022.findings-acl.177/}.

\bibitem[Masry et~al.(2023)Masry, Kavehzadeh, Do, Hoque, and Joty]{masry2023unichart}
Ahmed Masry, Parsa Kavehzadeh, Xuan~Long Do, Enamul Hoque, and Shafiq Joty.
\newblock Unichart: A universal vision-language pretrained model for chart comprehension and reasoning.
\newblock \emph{arXiv preprint arXiv:2305.14761}, 2023.

\bibitem[Mathew et~al.(2021{\natexlab{a}})Mathew, Bagal, Tito, Karatzas, Valveny, and Jawahar]{mathew2021infographicvqa}
Minesh Mathew, Viraj Bagal, Rubèn~Pérez Tito, Dimosthenis Karatzas, Ernest Valveny, and C.~V Jawahar.
\newblock Infographicvqa, 2021{\natexlab{a}}.
\newblock URL \url{https://arxiv.org/abs/2104.12756}.

\bibitem[Mathew et~al.(2021{\natexlab{b}})Mathew, Karatzas, and Jawahar]{mathew2021docvqadatasetvqadocument}
Minesh Mathew, Dimosthenis Karatzas, and C.~V. Jawahar.
\newblock Docvqa: A dataset for vqa on document images, 2021{\natexlab{b}}.
\newblock URL \url{https://arxiv.org/abs/2007.00398}.

\bibitem[Radford et~al.(2021)Radford, Kim, Hallacy, Ramesh, Goh, Agarwal, Sastry, Askell, Mishkin, Clark, Krueger, and Sutskever]{clip}
Alec Radford, Jong~Wook Kim, Chris Hallacy, Aditya Ramesh, Gabriel Goh, Sandhini Agarwal, Girish Sastry, Amanda Askell, Pamela Mishkin, Jack Clark, Gretchen Krueger, and Ilya Sutskever.
\newblock Learning transferable visual models from natural language supervision, 2021.
\newblock URL \url{https://arxiv.org/abs/2103.00020}.

\bibitem[Schuhmann et~al.(2022)Schuhmann, Beaumont, Vencu, Gordon, Wightman, Cherti, Coombes, Katta, Mullis, Wortsman, Schramowski, Kundurthy, Crowson, Schmidt, Kaczmarczyk, and Jitsev]{schuhmann2022laion}
Christoph Schuhmann, Romain Beaumont, Richard Vencu, Cade Gordon, Ross Wightman, Mehdi Cherti, Theo Coombes, Aarush Katta, Clayton Mullis, Mitchell Wortsman, Patrick Schramowski, Srivatsa Kundurthy, Katherine Crowson, Ludwig Schmidt, Robert Kaczmarczyk, and Jenia Jitsev.
\newblock Laion-5b: An open large-scale dataset for training next generation image-text models, 2022.
\newblock URL \url{https://arxiv.org/abs/2210.08402}.

\bibitem[Shazeer et~al.(2017)Shazeer, Mirhoseini, Maziarz, Davis, Le, Hinton, and Dean]{moe}
Noam Shazeer, Azalia Mirhoseini, Krzysztof Maziarz, Andy Davis, Quoc~V. Le, Geoffrey~E. Hinton, and Jeff Dean.
\newblock Outrageously large neural networks: The sparsely-gated mixture-of-experts layer.
\newblock In \emph{5th International Conference on Learning Representations, {ICLR} 2017, Toulon, France, April 24-26, 2017, Conference Track Proceedings}. OpenReview.net, 2017.
\newblock URL \url{https://openreview.net/forum?id=B1ckMDqlg}.

\bibitem[Shi et~al.(2024)Shi, Liu, Wang, Liao, Radhakrishnan, Huang, Yin, Sapra, Yacoob, Shi, Catanzaro, Tao, Kautz, Yu, and Liu]{shi2024eagleexploringdesignspace}
Min Shi, Fuxiao Liu, Shihao Wang, Shijia Liao, Subhashree Radhakrishnan, De-An Huang, Hongxu Yin, Karan Sapra, Yaser Yacoob, Humphrey Shi, Bryan Catanzaro, Andrew Tao, Jan Kautz, Zhiding Yu, and Guilin Liu.
\newblock Eagle: Exploring the design space for multimodal llms with mixture of encoders, 2024.
\newblock URL \url{https://arxiv.org/abs/2408.15998}.

\bibitem[Shi et~al.(2016)Shi, Caballero, Husz{\'a}r, Totz, Aitken, Bishop, Rueckert, and Wang]{shi2016real}
Wenzhe Shi, Jose Caballero, Ferenc Husz{\'a}r, Johannes Totz, Andrew~P Aitken, Rob Bishop, Daniel Rueckert, and Zehan Wang.
\newblock Real-time single image and video super-resolution using an efficient sub-pixel convolutional neural network.
\newblock In \emph{Proceedings of the IEEE conference on computer vision and pattern recognition}, pp.\  1874--1883, 2016.

\bibitem[Wang et~al.(2024)Wang, Lv, Yu, Hong, Qi, Wang, Ji, Yang, Zhao, Song, Xu, Xu, Li, Dong, Ding, and Tang]{cogvlm}
Weihan Wang, Qingsong Lv, Wenmeng Yu, Wenyi Hong, Ji~Qi, Yan Wang, Junhui Ji, Zhuoyi Yang, Lei Zhao, Xixuan Song, Jiazheng Xu, Bin Xu, Juanzi Li, Yuxiao Dong, Ming Ding, and Jie Tang.
\newblock Cogvlm: Visual expert for pretrained language models, 2024.
\newblock URL \url{https://arxiv.org/abs/2311.03079}.

\bibitem[Yue et~al.(2024)Yue, Ni, Zhang, Zheng, Liu, Zhang, Stevens, Jiang, Ren, Sun, Wei, Yu, Yuan, Sun, Yin, Zheng, Yang, Liu, Huang, Sun, Su, and Chen]{yue2024mmmumassivemultidisciplinemultimodal}
Xiang Yue, Yuansheng Ni, Kai Zhang, Tianyu Zheng, Ruoqi Liu, Ge~Zhang, Samuel Stevens, Dongfu Jiang, Weiming Ren, Yuxuan Sun, Cong Wei, Botao Yu, Ruibin Yuan, Renliang Sun, Ming Yin, Boyuan Zheng, Zhenzhu Yang, Yibo Liu, Wenhao Huang, Huan Sun, Yu~Su, and Wenhu Chen.
\newblock Mmmu: A massive multi-discipline multimodal understanding and reasoning benchmark for expert agi, 2024.
\newblock URL \url{https://arxiv.org/abs/2311.16502}.

\bibitem[Zhai et~al.(2023)Zhai, Mustafa, Kolesnikov, and Beyer]{siglip}
Xiaohua Zhai, Basil Mustafa, Alexander Kolesnikov, and Lucas Beyer.
\newblock Sigmoid loss for language image pre-training, 2023.
\newblock URL \url{https://arxiv.org/abs/2303.15343}.

\bibitem[Zhang et~al.(2024)Zhang, Zhang, Gu, Zhou, Lipka, Yang, and Sun]{zhang2024llavarenhancedvisualinstruction}
Yanzhe Zhang, Ruiyi Zhang, Jiuxiang Gu, Yufan Zhou, Nedim Lipka, Diyi Yang, and Tong Sun.
\newblock Llavar: Enhanced visual instruction tuning for text-rich image understanding, 2024.
\newblock URL \url{https://arxiv.org/abs/2306.17107}.

\bibitem[Zong et~al.(2024)Zong, Ma, Shen, Song, Shao, Jiang, Li, and Liu]{zong2024movaadaptingmixturevision}
Zhuofan Zong, Bingqi Ma, Dazhong Shen, Guanglu Song, Hao Shao, Dongzhi Jiang, Hongsheng Li, and Yu~Liu.
\newblock Mova: Adapting mixture of vision experts to multimodal context, 2024.
\newblock URL \url{https://arxiv.org/abs/2404.13046}.

\bibitem[Zoph et~al.(2022)Zoph, Bello, Kumar, Du, Huang, Dean, Shazeer, and Fedus]{zoph2022stmoedesigningstabletransferable}
Barret Zoph, Irwan Bello, Sameer Kumar, Nan Du, Yanping Huang, Jeff Dean, Noam Shazeer, and William Fedus.
\newblock St-moe: Designing stable and transferable sparse expert models, 2022.
\newblock URL \url{https://arxiv.org/abs/2202.08906}.

\end{thebibliography}




\end{document}